%% file: 00Main.tex
\documentclass[lettersize,journal]{IEEEtran}
\usepackage{amsmath,amsfonts}
\usepackage{algorithmic}
\usepackage{algorithm}
\usepackage{array}
\usepackage[caption=false,font=normalsize,labelfont=sf,textfont=sf]{subfig}
\usepackage{textcomp}
\usepackage{stfloats}
\usepackage{url}
\usepackage{verbatim}
\usepackage{graphicx}
\usepackage{cite}
\usepackage{tikz}
\usepackage{pgfplots} %引用包
\usepackage{multirow}
\usepackage{xspace}

\usepackage{scalerel}
\usetikzlibrary{svg.path}
\definecolor{orcidlogocol}{HTML}{A6CE39}
\tikzset{
    orcidlogo/.pic={
        \fill[orcidlogocol] svg{M256,128c0,70.7-57.3,128-128,128C57.3,256,0,198.7,0,128C0,57.3,57.3,0,128,0C198.7,0,256,57.3,256,128z};
        \fill[white] svg{M86.3,186.2H70.9V79.1h15.4v48.4V186.2z}
        svg{M108.9,79.1h41.6c39.6,0,57,28.3,57,53.6c0,27.5-21.5,53.6-56.8,53.6h-41.8V79.1z M124.3,172.4h24.5c34.9,0,42.9-26.5,42.9-39.7c0-21.5-13.7-39.7-43.7-39.7h-23.7V172.4z}
        svg{M88.7,56.8c0,5.5-4.5,10.1-10.1,10.1c-5.6,0-10.1-4.6-10.1-10.1c0-5.6,4.5-10.1,10.1-10.1C84.2,46.7,88.7,51.3,88.7,56.8z};
    }
}
\newcommand\orcidicon[1]{\href{https://orcid.org/#1}{\mbox{\scalerel*{
                \begin{tikzpicture}[yscale=-1,transform shape]
                \pic{orcidlogo};
                \end{tikzpicture}
            }{|}}}}
\usepackage[implicit=false]{hyperref}
\hypersetup{hidelinks,
	colorlinks=true,
	allcolors=black,
	pdfstartview=Fit,
	breaklinks=true}
 
\newcommand{\name}[0]{STN\xspace}
\usepackage{xcolor}

\begin{document}

\title{Siamese Transformer Networks for Few-shot\\ Image Classification} %Employing Gloabl and Local Features in 

\author{Weihao Jiang,  %riverfuture@hust.edu.cn
Shuoxi Zhang, %e-mail
Kun He${\textsuperscript{\orcidicon{0000-0001-7627-4604}}}$,~\IEEEmembership{Senior Member,~IEEE}
         % <-this % stops a space
\thanks{The authors are with School of Computer Science and Technology, Huazhong University of Scinece and Technology. The corresponding author is Kun He. Email: brooklet60@hust.edu.cn }
\thanks{This paper is supported by National Natural Science Foundation of China (U22B2017).
}% <-this % stops a space
\thanks{Manuscript received July 15, 2024; revised XXXX XX, XXXX.}}

\markboth{}%
{Shell \MakeLowercase{\textit{et al.}}: A Sample Article Using IEEEtran.cls for IEEE Journals}

\IEEEpubid{0000--0000/00\$00.00~\copyright~2024 IEEE}

\maketitle

\begin{abstract}
Humans exhibit remarkable proficiency in visual classification tasks, accurately recognizing and classifying new images with minimal examples. This ability is attributed to their capacity to focus on details and identify common features between previously seen and new images. In contrast, existing few-shot image classification methods often emphasize either global features or local features, with few studies considering the integration of both. To address this limitation, we propose a novel approach based on the Siamese Transformer Network (STN).
Our method employs two parallel branch networks utilizing the pre-trained Vision Transformer (ViT) architecture to extract global and local features, respectively. Specifically, we implement the ViT-Small network architecture and initialize the branch networks with pre-trained model parameters obtained through self-supervised learning. We apply the Euclidean distance measure to the global features and the Kullback–Leibler (KL) divergence measure to the local features. To integrate the two metrics, we first employ \(L_2\) normalization and then weight the normalized results to obtain the final similarity score. This strategy leverages the advantages of both global and local features while ensuring their complementary benefits.
During the training phase, we adopt a meta-learning approach to fine-tune the entire network. Our strategy effectively harnesses the potential of global and local features in few-shot image classification, circumventing the need for complex feature adaptation modules and enhancing the model's generalization ability. Extensive experiments demonstrate that our framework is simple yet effective, achieving superior performance compared to state-of-the-art baselines on four popular few-shot classification benchmarks in both 5-shot and 1-shot scenarios.

\end{abstract}

%-----------------------
\section{Introduction}
\input{01Introduction}

\section{Related Works}
\input{02RelatedWork}

\section{Methodology}
\input{03Method}

\section{Experiments}
 \input{04Experiments}

\section{Conclusion}
\input{05Conclusion}

%------------------------

\section*{Acknowledgments}
This work is supported by National Natural Science Foundation. %(U22B2017).

%-------------------references--------------------
\bibliographystyle{IEEEtran}
\bibliography{references}

\end{document}

%% file: 01Introduction.tex
Learning from just a few examples and applying that knowledge to diverse situations exemplifies human visual intelligence.
In recent years, leveraging deep learning techniques and large-scale labeled datasets has led to notable advancements in image recognition. 
Nevertheless, machine learning models have yet to match the adaptability of human visual cognition.
Humans effortlessly learn to identify new object classes with minimal examples, a cognitive feat enabling flexible adaptation of existing knowledge to new tasks. 
Inspired by this human capability, few-shot learning~\cite{siamese,one-shot,PrototypicalNetwork,hri,joc} addresses this challenge through knowledge transfer, employing a metric-based meta-learning approach~\cite{RelationNetwork,PrototypicalNetwork,Meta-Baseline} known for its simplicity and effectiveness. This approach significantly streamlines the application of deep learning in multimedia systems, garnering increasing attention in current literature.

\begin{figure}[!t]
\centering
\subfloat[Two images with similar global features%that look very similar
]{\includegraphics[width=2.8in]{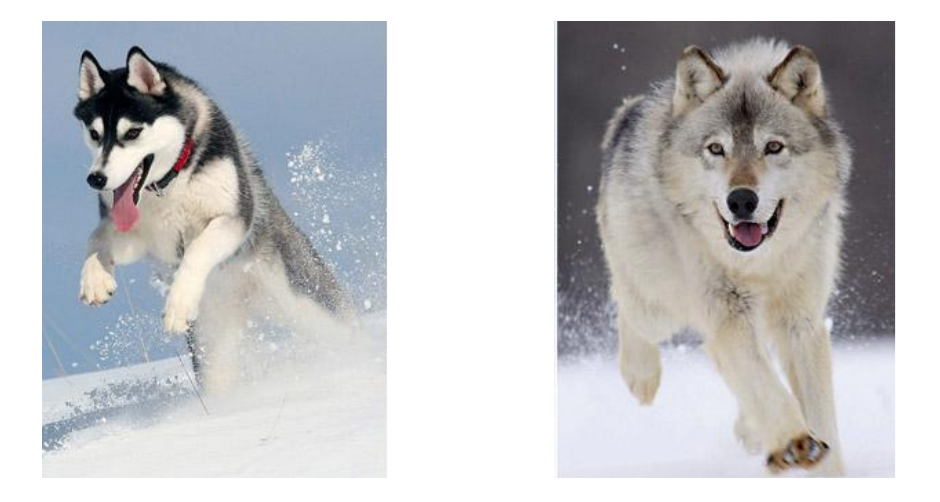}%
\label{fig_first_case}}
\hfil
\subfloat[Two images with similar details]{\includegraphics[width=2.8 in]{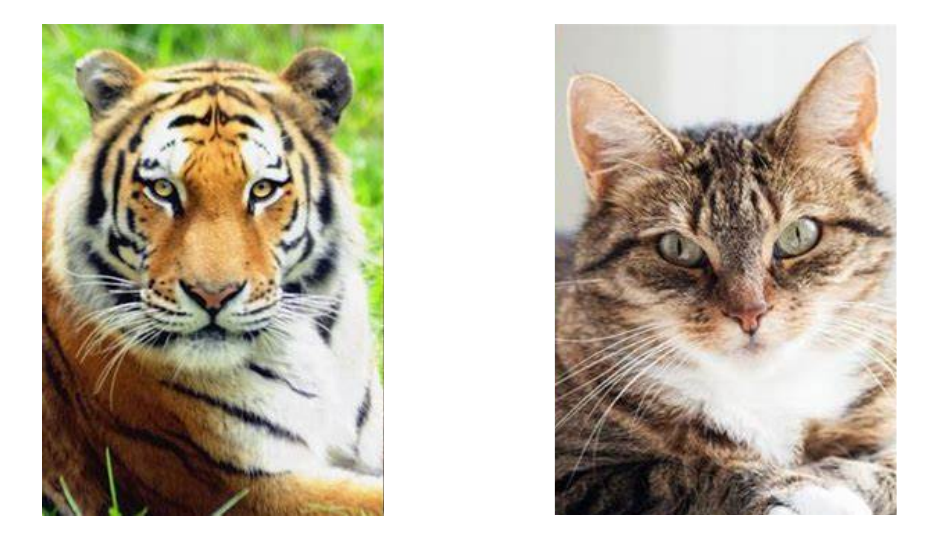}%
\label{fig_second_case}}
\caption{Illustration on limitations of either global feature-based methods or local feature-based methods. (a) When %a pair of
two images exhibit high similarity, global feature-based methods are susceptible to misjudgment. (b) %A pair of
Two images containing numerous similar details poses a challenge for local feature-based methods.}
\label{fig_sim}
\end{figure}

\IEEEpubidadjcol

The few-shot learning method based on metric learning primarily comprises two essential components: feature extraction from all query and support images, and the calculation of distances between query images and each support image, prototype, or class center using metrics. Feature extraction is a crucial step for achieving efficient measurement, laying the foundation for subsequent processing. Current feature extraction strategies can be divided into two main approaches, namely global feature-based~\cite{MatchingNetwork,PrototypicalNetwork,msml,deepemd,CTM,FEAT,MAML,ANIL,BOIL,LEO,dynamic,baseline++,RFS,P>M>F,HCTransformer} and local feature-based~\cite{covamnet,DN4,RelationNetwork,adm,deepemd,CAN,CTX,ATL-Net,RENet,FewTrue,cpea}. 
The global feature-based approach compresses the entire image into a unified feature vector, simplifying subsequent measurement process. To enhance the model's representational capacity, researchers focus on two optimization solutions: designing more effective network architectures~\cite{CTM,FEAT,RFS,P>M>F,HCTransformer} to extract more precise prototype features, or developing more accurate similarity evaluation methods~\cite{MatchingNetwork,PrototypicalNetwork,msml,deepemd} to ensure precise measurement results between different images.
In contrast, the local feature-based method decomposes each image into multiple local regions and extracts the corresponding features from each region. This approach not only emphasizes the use of local features for accurate measurement but also aims to understand the semantic associations between different images. By capturing detailed information within the image and establishing richer associations between different images, this method improves the robustness and accuracy of the measurement.

However, relying solely on either global or local features for measurement is prone to introducing bias. As shown in Figure~\ref{fig_first_case}, the comparative images of a dog and a wolf are so similar overall that depending exclusively on global characteristics can lead to misjudgment. Similarly, Figure~\ref{fig_second_case} illustrates a pair of contrasting images of a tiger and a cat, which share numerous detailed similarities, potentially causing mistaken identification if only local features are used for comparison. Therefore, it is logical to combine global and local feature-based measurements to achieve more accurate results than relying on either one alone.

Although global feature-based and local feature-based methods have been thoroughly studied and have demonstrated advanced performance, a critical area has not received substantial attention: integrating measurements based on global features with those based on local features. While some studies~\cite{mixer,cpea} have attempted to combine these two types of features, they generally use global features to enhance local features rather than treating them as separate and equally important components.
Global features are favored for their inherent invariance properties but are susceptible to disruption by image noise, especially in the presence of complex backgrounds. Conversely, local features excel at capturing detailed image information and are robust to noise and minor changes, yet they face challenges with alignment. The complementary strengths of global and local features suggest that their combined utilization could achieve synergistic advantages and further improve model performance.
However, simply measuring based on fused features cannot fully exploit the potential of each type of feature. In contrast, fusing the measurement results from different features allows for more flexibility. This approach can employ distinct and more appropriate measurement methods tailored to the characteristics of each feature, thereby fully leveraging the strengths of both global and local features.

To fully leverage the complementary strengths of global and local features, we design a novel model called the Siamese Transformer Network (\name). The \name model consists of two distinct branches, each responsible for extracting the global features and the local feature sets, respectively.
Specifically, the first branch network is dedicated to extracting the global features of the support and query images, while the second branch network focuses on extracting the local feature sets for the two sets of images. 
We carefully design different metrics to measure the similarity between the global features and the local feature sets of the query and support images. For the global features, we employ a regular Bregman divergence~\cite{PrototypicalNetwork} to compute the similarity between samples. For the local feature sets, we adopt an asymmetric distribution metric.
To synthesize the two measurements, we first perform $L_2$ normalization to ensure that the measurements are of the same scale. We then weight the two normalized results to obtain the final metric score, which reflects the overall similarity between the query image and the support image.
Finally, based on this aggregate metric score, we assign a corresponding label to the query image. This method not only fully utilizes the information from both global and local features but also improves the accuracy and reliability of the measurement results.

Our main contributions can be summarized as follows:

$\bullet$ We present an innovative approach that ingeniously integrates the dual perspectives of global and local features, transcending the limitations of existing metric-based methods that narrowly focus on a single feature dimension. This methodology captures the holistic structure of images while meticulously analyzing intricate textures, thereby charting a new course in the field of image analysis and significantly enhancing the comprehensiveness and precision of feature extraction.

$\bullet$ We introduce \name, a novel approach that combines measurements based on both global and local image features without requiring other complex designs, thereby enhancing model generalization.

$\bullet$ Extensive experiments conducted on four prominent benchmark datasets demonstrate that \name outperforms state-of-the-art methods. Comprehensive comparisons further highlight the superiority of our approach.

The remainder of this paper is organized as follows: Section II provides a summary of related work. Section III elaborates on our approach. The experiments conducted are detailed in Section IV. Section V concludes this work.

%% file: 02RelatedWork.tex
Here, we briefly review few-shot learning methods and categorize existing methods into two groups based on the  characteristics of adopted features. We then introduce Siamese neural networks and analyze how our approach differs.

\subsection{Global Feature-Based Methods} 
Global feature-based methods utilize global features, representing an image through a singular global feature. Metric-based few-shot learning (FSL) methods embed both support and query images into a shared feature space, classifying query images by measuring their distance or similarity to support samples. These approaches primarily focus on enhancing similarity measures and improving feature extraction techniques. Several models have been developed employing various metric methods, including Matching Networks~\cite{MatchingNetwork}, Prototypical Networks~\cite{PrototypicalNetwork}, MSML~\cite{msml}, and DeepEMD~\cite{deepemd}.
To obtain more distinctive features, researchers have begun to adopt a holistic approach to the entire task, developing external modules to enhance feature representation. For instance, CTM~\cite{CTM} introduces a Category Traversal Module that traverses the entire support set simultaneously, identifying task-relevant features based on intra-class commonality and inter-class uniqueness within the feature space. Similarly, FEAT~\cite{FEAT} proposes a novel method to adapt instance-level global features to the target classification task using a set-to-set function, thereby generating task-specific and discriminative global features.
Optimization-based methods aim to learn effective model initializations, facilitating rapid optimization when confronted with new tasks. A prominent example in this category is MAML~\cite{MAML}, which achieves optimal initialization parameters through both internal and external training during the meta-training phase. MAML has inspired numerous subsequent efforts, including ANIL~\cite{ANIL}, BOIL~\cite{BOIL}, and LEO~\cite{LEO}.
Transfer-based methods initially employ a straightforward scheme to train a classification model on the complete training set. After training, the classification head is removed, retaining only the feature extraction component, and a new classifier is trained based on the support set from the testing data. Notable examples in this area include Dynamic Classifier~\cite{dynamic}, Baseline++\cite{baseline++}, and RFS\cite{RFS}.

Recently, researchers have begun incorporating Vision Transformers (ViT) into few-shot learning scenarios. The PMF method~\cite{P>M>F} first utilizes a pre-trained Transformer model with external unsupervised data. It then simulates few-shot tasks for meta-training using base categories and subsequently fine-tunes the model with limited labeled data from test tasks.
HCTransformer~\cite{HCTransformer} employs hierarchically cascaded Transformers as a robust meta-feature extractor for few-shot learning.
\subsection{Local Feature-Based Methods}  
In contrast, local feature-based methods focus on local features by decomposing an image into a series of features that represent different local regions. Each local region often contains distinct content and carries unique semantic meaning, leading researchers to explore these methods in two main directions. On one hand, there is ongoing work to design more fine-grained local feature measures that capture subtle differences between features. On the other hand, researchers are seeking strategies to achieve explicit semantic alignment between images.

In the first aspect, several methods have been proposed. For example, CovaMNet~\cite{covamnet} introduces a deep covariance metric to measure the consistency of distributions between query samples and support samples. DN4~\cite{DN4} calculates instance-to-class similarity between a query image and support images using \(k\)-nearest neighbors. RelationNet~\cite{RelationNetwork} employs a learnable deep metric network to assess the relationships between images. Additionally, ADM~\cite{adm} proposes a novel asymmetric distribution measure network specifically for few-shot learning.
In the second direction, researchers achieve semantic alignment by designing learnable modules that guide the model to focus on target areas within the image. For instance, SAML~\cite{saml} utilizes an attention mechanism to diminish the influence of semantically unrelated areas. DeepEMD~\cite{deepemd} promotes semantic correspondence between images by minimizing the Earth Mover’s Distance. CAN~\cite{CAN} generates cross-attention maps for each pair of support feature maps and query sample feature maps to emphasize target object regions, thereby enhancing the discriminative nature of the extracted features. 
CTX~\cite{CTX} learns spatial and semantic alignment between CNN-extracted query and support features using a Transformer-style attention mechanism. ATL-Net~\cite{ATL-Net} introduces episodic attention calculated through a local relation map between the query image and the support set, adaptively selecting important local patches across the entire task. RENet~\cite{RENet} combines self-correlational representations within each image with cross-correlational attention modules between images to learn relational embeddings. Lastly, FewTURE~\cite{FewTrue} adopts a fully transformer-based architecture, learning token importance weights through online optimization during inference. CPEA~\cite{cpea} also employs a pre-trained ViT model, integrating patch embeddings with class-aware embeddings to enhance class relevance.

Existing methods typically rely solely on either global or local features for similarity measurement, which limits their ability to capture the synergistic effects of utilizing both types of features. While some studies incorporate both global and local features, they primarily use global features to enhance the semantic information of local features~\cite{cpea, mixer}, ultimately performing measurements based exclusively on local features. This reliance restricts the accuracy of the measurement results.
To address this limitation, we propose a Siamese Transformer Network that simultaneously extracts both global and local image features. The network measures these two distinct feature sets separately using different approaches, then fuses the resulting measurements to produce a final similarity score.

\subsection{Siamese Neural Networks}  
Siamese neural networks were first proposed by Bromley and LeCun in the early 1990s to address signature verification as an image matching problem~\cite{siamese}. 
These networks consist of two branches, each receiving an input, mapping it to a high-dimensional feature space, and outputting the corresponding representation. By calculating the distance between the two representations —-- commonly using metrics like Euclidean distance —-- the model can assess the similarity between the inputs. The branch networks can be composed of convolutional neural networks (CNNs) or recurrent neural networks (RNNs), with weights optimized using energy functions or classification losses~\cite{2015siamese}.

In a narrow sense, Siamese neural networks are defined as having two identical neural networks with shared weights~\cite{siamese}. A generalized version, known as a ``Pseudo-siamese network'', can consist of any two neural networks~\cite{siam}. The primary distinction is that in Pseudo-Siamese networks, the weights of the two branches are not shared, resulting in a true dual-branch architecture. Each branch represents a different mapping function, meaning the feature extraction structures differ, with distinct weights or network layers. Consequently, this architecture has nearly twice as many training parameters as the traditional Siamese network, providing greater flexibility.

In this paper, we employ a Vision Transformer as the backbone for our dual-branch network, ensuring no parameter sharing between the two branches. After extracting the global and local features from the input image, we compute similarity scores based on each type of feature, then normalize and weight these measures to obtain a final similarity score.

%% file: 03Method.tex
In this section, we first provide definition on the few-shot image classification task. Next, we outline the framework of our proposed method. Following that, we provide a detailed description of our Siamese Transformer Network module and explain the classification method based on two metrics of global features and local features.

\subsection{Problem Definition}
Few-shot image classification is primarily concerned with the $N$-way $K$-shot problem, where $N$ denotes the number of categories and $K$ represents the number of instances within each category. Typically, $K$ is relatively small like 1 or 5.

Datasets designed for few-shot learning typically comprise three distinct parts: training set $\mathcal{D}_{\rm{train}}$, validation set $\mathcal{D}_{\rm{val}}$, and test set $\mathcal{D}_{\rm{test}}$. Notably, they feature non-overlapping categories, indicating that images in the test set are entirely unseen during training and validation. This lack of overlap poses a significant challenge for few-shot learning. Typically, all the three datasets contain numerous categories and examples. To emulate the conditions of few-shot learning, researchers adopt an episode training mechanism~\cite{MatchingNetwork}. In this mechanism, episodes are randomly sampled from the datasets, each consisting of a support set 
$S = \left \{ (x_i,y_i) \right \} _{i=1}^{NK}$ and a query set 
$Q = \left \{ (\tilde{x}_i ,\tilde{y}_i) \right \} _{i=1}^{NT}$, where $T$ is the number of query examples contained in each class. Both the $S$ and $Q$ have identical labels, but their samples do not overlap 
($S \cap Q = \emptyset$). The support set contains a sparse selection of labeled samples, acting as the few-shot training data, while the query set is used for evaluation.

During the training phase, a large number of episodes sampled from the training set are utilized to update the model parameters until reaching convergence. For validation and testing, episodes from the validation and test sets are employed to ensure that the test conditions mirror real-world scenarios.

%=====================
\begin{figure*}[t]
\begin{center}
  \includegraphics[width=1\linewidth]{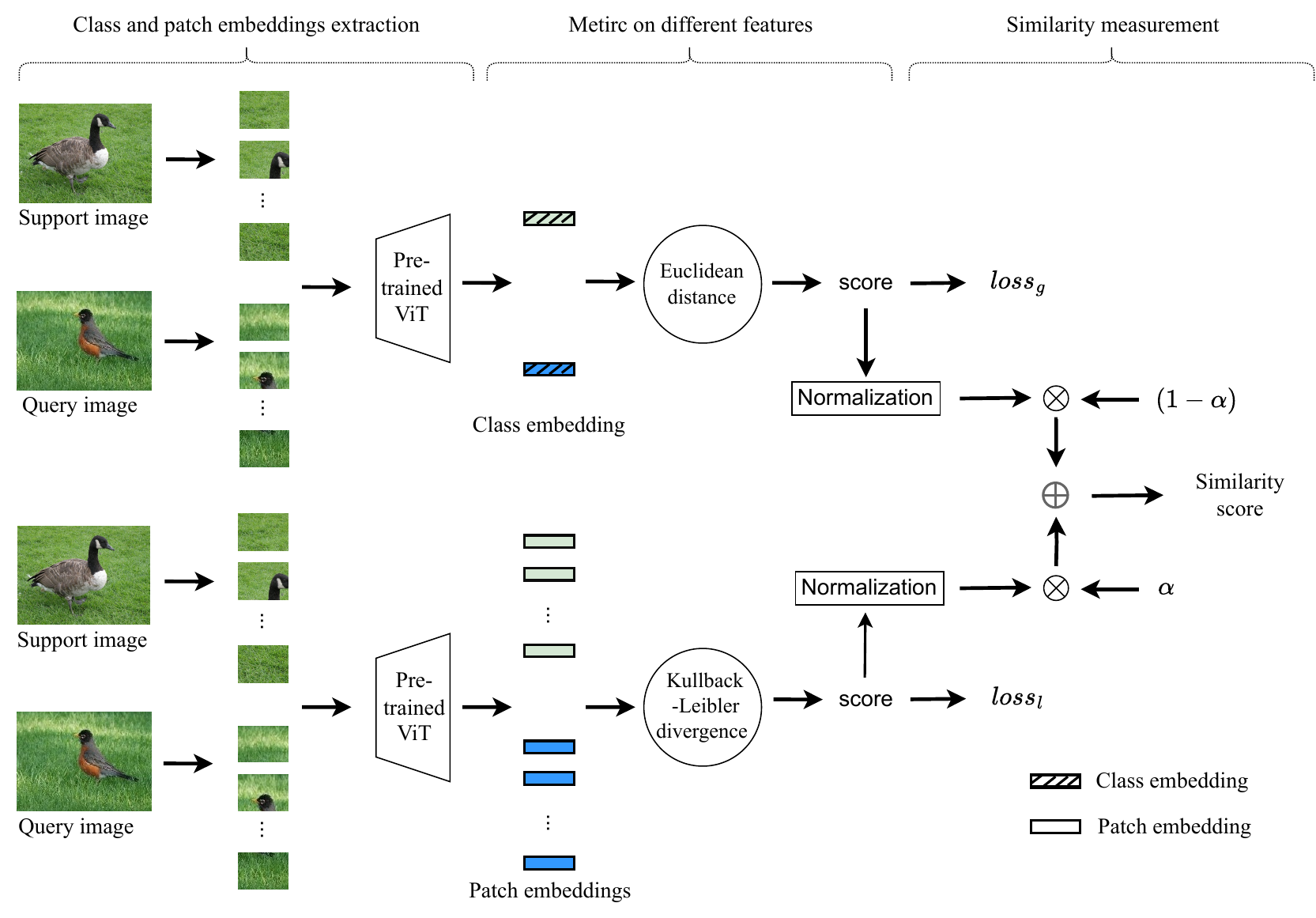}
\end{center}
   \caption{
   The processing pipeline of \name involves several key steps. Support and query images are first divided into patches and then encoded using a pre-trained Vision Transformer (ViT). Class embeddings capture the global features, while patch embeddings represent the local features. Using these two types of features, Euclidean distance and KL divergence are employed for measurement. Independent branch network optimization is performed based on their respective evaluation results. Finally, the scores from both measurements are normalized and weighted to generate the final similarity scores.
   }
\label{fig:framework}
\end{figure*}

\subsection{The Proposed Framework} 
The framework of our \name model, as illustrated in Figure~\ref{fig:framework}, leverages the Vision Transformer (ViT) as the feature extractor. Each input image is first partitioned into non-overlapping patches and then encoded using the pre-trained ViT~\cite{ViT}, resulting in both class embeddings and patch embeddings. The class embeddings capture global image information, while the patch embeddings focus on local details specific to each patch.

To conduct a comprehensive analysis of the input images (including both query and support images), we design a dual-branch network architecture. The first branch is dedicated to extracting global features, while the second branch focuses on extracting local features.
To measure similarity between images based on these features, two distinct strategies are employed. Specifically, Euclidean distance serves as the similarity metric for global features, while KL divergence assesses the differences between images based on local features. This approach leverages the strengths of each feature type, providing a more thorough evaluation of similarity by combining measurements from both branches.
To integrate the similarity information from global and local features, we first normalize the measurement results using the \(L_2\) normalization. The final similarity score is then obtained through a weighted summation of the normalized values. In this process, the weighted coefficients act as the sole hyperparameter, eliminating the need for complex feature adaptation modules.

In contrast to traditional methods, the proposed approach introduces several key innovations. First, it utilizes both global and local features to assess image similarity and effectively combines the outcomes of these complementary measures. Second, it eliminates the need for complex feature adaptation modules by employing a feature extraction model that generates diverse feature perspectives, thereby enhancing the model's generalization capability.
Prior studies addressing image similarity measurement have either focused on adjusting global features to become more task-specific or used them to enhance the semantic information of local features. However, these methods typically rely on a single type of feature for comparison, failing to fully exploit the advantages of combining global and local characteristics. Additionally, these approaches often depend on supplementary network modules for parameter optimization and learning, which can increase model complexity and negatively impact the generalization ability.

The proposed approach eliminates the need for additional network modules by employing a concise strategy for integrating global and local features. Specifically, similarity measures based on global and local features are calculated independently and then synthesized using a single weighted parameter. This streamlined method reduces model complexity and enhances generalization ability by minimizing the introduction of excessive learnable parameters outside the feature extractor.
Through this weighted fusion mechanism, the approach effectively combines global contextual information with detailed localized semantics, resulting in improved performance on the image similarity measurement task.

\subsection{Siamese Transformer Networks} 
Our Siamese Transformer Network consists of two branches, each utilizing a Vision Transformer as the backbone.

In a typical few-shot task, the support set usually consists of \(N\) classes, commonly set to five, following the standard few-shot learning setup. Each support category contains \(K\) shots, and the average of the embedding values is used as the prototype embeddings~\cite{PrototypicalNetwork}. After the ViT encoding process, we obtain five sets of combinations of class embeddings and patch embeddings:
\begin{align}\label{e1}
    S_i &= [class_{i}^s, patch_{i,1}^s, patch_{i,2}^s, \ldots, patch_{i,M}^s],\\
    &\quad i \in [1,5], \nonumber
\end{align}
where \(M\) is the number of patches.

For a query image, we have a combination of class embedding and patch embeddings:
\begin{align}\label{e2}
    Q &= [class^q, patch_{1}^q, patch_{2}^q, \ldots, patch_{M}^q],
\end{align}
where \(M\) is the number of patches. Here, \(class\) represents the class embedding that captures the global information of the image, while \(patch\) denotes the patch embeddings that capture features from local regions of the image.

Global features are valued for their invariance but can be easily affected by image noise, particularly in complex backgrounds. In contrast, local features excel at capturing detailed image information and are robust to noise and small variations, although they often struggle with alignment issues.
To address these challenges, the proposed approach employs different similarity measures based on these two types of features and subsequently fuses the measurement results. This strategy leverages the complementary advantages of both feature types, thereby enhancing the overall performance of the model.

For the first branch network, the class embedding is obtained as the global feature:
\begin{align}\label{e3}
    S_g  = \frac{1}{K}  \sum_{i=1}^{K} class_{i}^s
\end{align}

\begin{align}\label{e4}
    Q_g  = class^q
\end{align}

For the second branch network, the patch embeddings are obtained as the local features:
\begin{align}\label{e5}
    S_l  = [patch_{i,1}^s, patch_{i,2}^s, \ldots, patch_{i,M}^s],
\end{align}

\begin{align}\label{e6}
    Q_l  = [patch_{1}^q,patch_{2}^q,...,patch_{M}^q],
\end{align}
where $M$ is the number of patches.

Based on the global features, we utilize squared Euclidean distance as the measurement. This metric is widely used as a similarity measure across various fields~\cite{PrototypicalNetwork, FEAT} and has demonstrated effectiveness in practice.
\begin{align}\label{e7}
    D_{Ed}(Q,S^{n}) &= \left \| Q_g-S^{n}_g \right \| ^{2} ,\\
    &n\in [1,5], \nonumber
\end{align}
where \(S^{n}_g\) represents the global feature of the \(n\)-th support class.

Based on the local features, we employ an asymmetric Kullback–Leibler (KL) divergence measure~\cite{adm} to align the distribution of a query with that of a support class, capturing global distribution-level asymmetric relations. The KL divergence assumes that the distribution of local features extracted from an image or a support class follows a multivariate Gaussian distribution. Specifically, the distribution of a query image can be represented as \(Q = \mathcal{N}(\mu_Q, \Sigma_Q)\), while the distribution for a support class can be expressed as \(S = \mathcal{N}(\mu_S, \Sigma_S)\), where \(\mu \in \mathbb{R}^c\) and \(\Sigma \in \mathbb{R}^{c \times c}\) denote the mean vector and covariance matrix of the respective distributions. Thus, the KL divergence~\cite{kl} between \(Q\) and \(S\) can be defined as:
\begin{align}\label{e8}
    D_{KL}(Q||S) = \frac{1}{2}(trace(\Sigma^{-1}_S \Sigma_Q) + ln\left(\frac{det \Sigma_S}{det \Sigma_Q} \right) \\ 
              + (\mu_S-\mu_Q)^T \Sigma^{-1}_S(\mu_S-\mu_Q)-c), \nonumber
\end{align}
where \(\text{trace}(\cdot)\) represents the trace operation of a matrix, \(\ln(\cdot)\) denotes the logarithm with base \(e\), and \(\det\) indicates the determinant of a square matrix. Equation~\eqref{e8} considers both the mean and covariance to calculate the distance between two distributions. A significant advantage of using this equation is its ability to naturally capture the asymmetric relationship between a query image and a support class through the local feature set. This encourages the query images to be closer to the corresponding true class during network training~\cite{adm}.

Finally, we combine the distribution-based KL divergence measure with the Euclidean distance measure to simultaneously capture both global and local relations.

\subsection{Classification With an Additive Fusion Strategy} 
As two distinct types of metrics have been calculated --- local-level relations quantified by the Kullback-Leibler divergence measure, and global-level relations produced by the Euclidean distance measure --- a fusion strategy must be designed to integrate these two components. To address this, the proposed approach adopts a hyperparametric weight $\alpha$ to implement the fusion.
It is worth noting that since the KL divergence indicates dissimilarity rather than similarity, the negative of this divergence is used to obtain a similarity measure, akin to the Euclidean distance. Specifically, the final fused similarity between a query $Q$ and a class $S$ can be defined as follows:
\begin{align}\label{e9}
    D(Q,S) = -\alpha \cdot D_{KL}(Q||S) - (1-\alpha)\cdot D_{Ed}(Q,S)
\end{align}

For a 5-way 1-shot task and a specific query \(Q\), the output from each branch produces a 5-dimensional similarity vector. Subsequently, \(L_2\) normalization is applied to balance the scales of the two similarity components. Following this, we use an additive fusion strategy, as described in Equation~\eqref{e9}, to combine the two metrics. This process results in an integrated 5-dimensional similarity vector derived from two distinct feature-based metrics. Finally, a non-parametric nearest neighbor classifier is employed to obtain the final classification results.

\subsection{Training and Inference}
In the training process, we adopt a strategy of separation and parallelism. Each branch network computes its own evaluation results based on different feature measures and optimizes independently using the cross-entropy loss function. 
For the first branch network, the Euclidean distance measurement is adopted based on global features. The probability of the $i$-th query image belonging to the $n$-th support class can then be calculated as follows:
\begin{align}\label{e10}
    p^g_{ni}=\frac{exp(-D_{Ed}(Q^i,S^n)}{ {\textstyle \sum_{{n}'=1}^{N}}exp(-D_{Ed}(Q^i,S^{{n}'})) } ,
\end{align}
for the second branch network, KL divergence is adopted based on local features, and the probability is as follows:
\begin{align}\label{e11}
    p^l_{ni}=\frac{exp(-D_{KL}(Q^i||S^n))}{ {\textstyle \sum_{{n}'=1}^{N}}exp(-D_{KL}(Q^i||S^{{n}'})) } .
\end{align}
For a given episode, the loss function of the first branch network can be formulated as follows:
\begin{align}\label{e12}
    Loss_g = -\frac{1}{NQ} \sum_{i=1}^{NQ} \sum_{n=1}^{N}\mathrm {\mathit{I} }(y^{(Q)}_i=n) \mathrm{log}p^g_{ni},
\end{align}
and the loss function of the second branch network can be formulated as follows:
\begin{align}\label{e13}
    Loss_l = -\frac{1}{NQ} \sum_{i=1}^{NQ} \sum_{n=1}^{N}\mathrm {\mathit{I} }(y^{(Q)}_i=n) \mathrm{log}p^l_{ni},
\end{align}
where $y^{(Q)}_i$ denotes the label of the $i$-th query image and $\mathrm {\mathit{I}}(\cdot)$ is an indicator function that equals one if its  arguments are true and zero otherwise. 
In the first branch network, all the learnable weights are fine-tuned by minimizing the loss function defined in Equation~\eqref{e12}.
Correspondingly, in the second branch network, all the learnable weights are adjusted using the minimization of the loss function in Equation~\eqref{e13}, where a randomly selected sample of training episodes is utilized.

\textbf{Inference}. Given an episode sampled from the unseen test classes, the probability of a query image belonging to each class can be calculated according to Equation~\eqref{e9}. Then, we assign the label of the class with the maximum probability to the corresponding query image.
It is worth noting that, once fine-tuned on the training classes, the method does not require any adjustments when generalizing to unseen test classes. This is in contrast to FewTURE~\cite{FewTrue}, which needs all images of an episode's support set together with their labels to learn the importance of each individual patch token via online optimization at inference time. As a result, the proposed approach is much simpler than FewTURE in terms of inference.

%% file: 04Experiments.tex
This section first details the experimental settings. Subsequently, a comparison with competing methods on benchmark datasets is presented. Finally, an ablation study of the proposed framework is conducted.
\subsection{Datasets} 
In the standard few-shot classification task, four popular benchmark datasets are commonly used: $mini$ImageNet~\cite{MiniImageNet}, $tiered$ImageNet~\cite{tieredImageNet}, CIFAR-FS~\cite{cifar-fs}, and FC100~\cite{fc100}.

The $mini$ImageNet dataset comprises 100 categories sampled from ILSVRC-2012~\cite{ImageNet}, with each category containing 600 images, totaling 60,000 images. In the standard setting, the dataset is randomly partitioned into training, validation, and testing sets, consisting of 64, 16, and 20 categories, respectively.

The $tiered$ImageNet dataset, also sourced from ILSVRC-2012, features a larger scale of data. It consists of 34 super-categories, divided into training, validation, and testing sets, containing 20, 6, and 8 super-categories, respectively. In total, the dataset includes 608 categories, with 351, 97, and 160 categories in each partitioned set.

Both CIFAR-FS and FC100 are derived from CIFAR-100, which consists of 100 classes with 600 images per class. These datasets feature small-resolution images, each measuring 32 × 32 pixels. Specifically, CIFAR-FS is randomly divided into 64 training classes, 16 validation classes, and 20 testing classes. In contrast, FC100 includes 100 classes sourced from 36 super-classes in CIFAR-100, which are organized into 12 training super-classes (60 classes), 4 validation super-classes (20 classes), and 4 testing super-classes (20 classes).

\subsection{Implementation Details} 
Motivated by the scalability and effectiveness of pre-training techniques, we employ a Masked Image Modeling (MIM)-pretrained Vision Transformer as the backbone of our model. Specifically, we utilize the ViT-Small architecture with a patch size of 16.
During the pre-training phase, we adopt the same strategy as outlined in \cite{FewTrue} to pretrain our ViT-Small backbones, adhering closely to the hyperparameter settings reported in their work. 
In the meta-training phase, we employ the AdamW optimizer with default settings. The initial learning rate is set to \(1 \times 10^{-5}\) and decays to \(1 \times 10^{-6}\) following a cosine learning rate schedule.
For the $mini$ImageNet and $tiered$ImageNet datasets, we resize the images to \(224 \times 224\) pixels and train for 100 epochs, with each epoch consisting of 200 episodes. Similarly, for the CIFAR-FS and FC100 datasets, we resize the images to \(224 \times 224\) pixels and train for 90 epochs, with 200 episodes per epoch. During fine-tuning, we implement an episodic training mechanism. To ensure consistency with prior studies, we utilize the validation set to select the best-performing models.
Additionally, we apply standard data augmentation techniques, including random resizing and horizontal flipping.

During the meta-testing phase, we randomly sample 1000 tasks, each containing 15 query images per class. We report the mean accuracy along with the corresponding $95\%$ confidence interval.

\begin{table*}[t]
\caption{
Comparisons on 5-way 1-shot and 5-way 5-shot with 95$\%$ confidence intervals on $mini$Imagenet and $tiered$Imagenet. } 
\label{tab:miniImage}
\begin{center}
% \scalebox{1.0}{
    \begin{tabular}{l|cl|cc|cc}
    \hline
    % \multirow{2}{*}{Model} &\multirow{2}{*}{Backbone} & \multirow{2}{*}{$\approx$Params} & \multicolumn{2}{c|}
    \multirow{2}{*}{Model} &\multirow{2}{*}{Backbone} & \multirow{2}{*}{Venue} & \multicolumn{2}{c|}{$mini$ImageNet} & \multicolumn{2}{c}{$tiered$ImageNet} \\
                           &                  &              &1-shot& 5-shot & 1-shot& 5-shot \\ 
    \hline
    \hline
    ProtoNet~\cite{PrototypicalNetwork}  & ResNet-12 & NeurIPS 2017 & 62.29±0.33  & 79.46±0.48 & 68.25±0.23  & 84.01±0.56   \\
    FEAT~\cite{FEAT} & ResNet-12 & CVPR 2020 &66.78±0.20 &82.05±0.14 &70.80±0.23 &84.79±0.16 \\
    CAN~\cite{CAN} & ResNet-12 & NeurIPS 2019 & 63.85±0.48  & 79.44±0.34 &69.89±0.51 &84.23±0.37 \\
    CTM~\cite{CTM} & ResNet-18 & CVPR 2019& 64.12±0.82 & 80.51±0.13 &68.41±0.39 &84.28±1.73 \\
    ReNet~\cite{RENet} & ResNet-12 & ICCV 2021& 67.60±0.44  & 82.58 ±0.30 &71.61±0.51 &85.28±0.35  \\
    DeepEMD~\cite{deepemd} & ResNet-12 &CVPR 2020 &65.91±0.82 &82.41±0.56 &71.16±0.87 &86.03±0.58 \\
    IEPT~\cite{IEPT} &ResNet-12 &ICLR 2021 & 67.05±0.44 & 82.90±0.30 & 72.24±0.50 & 86.73±0.34 \\
    MELR~\cite{MELR} &ResNet-12 &ICLR 2021 & 67.40±0.43 & 83.40±0.28 & 72.14±0.51 & 87.01±0.35 \\
    FRN~\cite{FRN} &ResNet-12 &CVPR 2021 & 66.45±0.19 & 82.83±0.13 & 72.06±0.22 & 86.89±0.14 \\
    CG~\cite{CG/CNL} &ResNet-12 &AAAI 2021 & 67.02±0.20 & 82.32±0.14 & 71.66±0.23 & 85.50±0.15 \\
    DMF~\cite{DMF} &ResNet-12 &CVPR 2021 & 67.76±0.46 & 82.71±0.31 & 71.89±0.52 & 85.96±0.35 \\
    InfoPatch~\cite{InfoPatch} &ResNet-12 &AAAI 2021 & 67.67±0.45 & 82.44±0.31 & - & -\\
    BML~\cite{BML} &ResNet-12 &ICCV 2021 & 67.04±0.63 & 83.63±0.29 & 68.99±0.50 & 85.49±0.34 \\
    CNL~\cite{CG/CNL} &ResNet-12 &AAAI 2021 & 67.96±0.98 & 83.36±0.51 & 73.42±0.95 & 87.72±0.75 \\
    Meta-NVG~\cite{Mata-NVG} &ResNet-12 &ICCV 2021 & 67.14±0.80 & 83.82±0.51 & 74.58±0.88 & 86.73±0.61 \\ 
    PAL~\cite{PAL} &ResNet-12 & ICCV 2021 & 69.37±0.64 & 84.40±0.44 & 72.25±0.72 & 86.95±0.47 \\
    COSOC~\cite{COSOC} &ResNet-12 & NeurIPS 2021 & 69.28±0.49 & 85.16±0.42 & 73.57±0.43 & 87.57±0.10 \\
    Meta DeepBDC~\cite{Meta-DeepBDC} &ResNet-12 & CVPR 2022 & 67.34±0.43 & 84.46±0.28 & 72.34±0.49 & 87.31±0.32 \\
    QSFormer~\cite{qsformer} &ResNet-12 & TCSVT 2023 & 65.24±0.28 & 79.96±0.20 & 72.47±0.31 & 85.43±0.22 \\
    LastShot~\cite{lastshot} &ResNet-12 & TPAMI 2024 & 67.35±0.20 & 82.58±0.14 & 72.43±0.23 & 85.82±0.16 \\
    \hline
    LEO~\cite{LEO} & WRN-28-10 & ICLR 2019 & 61.76±0.08 & 77.59±0.12 & 66.33±0.05 & 81.44±0.09 \\
    CC+rot~\cite{CC-rot} & WRN-28-10 & ICCV 2019 & 62.93±0.45 & 79.87±0.33 & 70.53±0.51 & 84.98±0.36 \\
    FEAT~\cite{FEAT} & WRN-28-10 & CVPR 2020 & 65.10±0.20 & 81.11±0.14 & 70.41±0.23 & 84.38±0.16 \\
    % PSST~\cite{PSST} & WRN-28-10 & 36.5 M & 64.16±0.44 & 80.64±0.32 & - & - \\
    MetaQDA~\cite{MetaQDA} & WRN-28-10 & ICCV 2021 & 67.83±0.64 & 84.28±0.69 & 74.33±0.65 & 89.56±0.79 \\ 
    OM~\cite{OM} & WRN-28-10 &  ICCV 2021 & 66.78±0.30 & 85.29±0.41 & 71.54±0.29 & 87.79±0.46 \\
    \hline
    FewTURE~\cite{FewTrue}  & ViT-Small & NeurIPS 2022 & 68.02±0.88 &  84.51±0.53 & 72.96±0.92 &  86.43±0.67\\
    % FewTURE~\cite{FewTrue}  & Swin-Tiny & 29 M & 72.40±0.78 &  86.38±0.49 & 76.32±0.87 & 89.96±0.55\\
    CPEA~\cite{cpea} & ViT-Small & ICCV 2023 & 71.97±0.65 & 87.06±0.38 & 76.93±0.70 & 90.12±0.45\\
    \hline
    % IMAformer (ours) & ViT-Small & 22 M & \emph{70.25±0.61}  &  \emph{86.48±0.44}  & \emph{75.15±0.67}  &  \emph{88.44±0.65} \\    % two layers       
    % IMAformer (ours) & ViT-Small & 22 M & \emph{78.48±0.39}  &  \emph{89.05±0.28}  & \emph{77.70±0.67}  &  \emph{90.68±0.65} \\
     \name (ours) & ViT-Small & - & \bf72.04±0.62  & \bf88.00±0.37  & \bf77.10±0.74  & \bf90.71±0.42 \\
    \hline
    \end{tabular}
\end{center}

\end{table*}

\begin{table*}[h!]
\caption{
Comparisons on 5-way 1-shot and 5-way 5-shot  with 95$\%$ confidence intervals on CIFAR-FS and FC100.}
\label{tab:cifar}
\begin{center}
% \scalebox{1.0}{
    \begin{tabular}{l|cl|cc|cc}
    \hline
    % \cline{1-1} \cline{3-5}
    \multirow{2}{*}{Model} &\multirow{2}{*}{Backbone} & \multirow{2}{*}{Venue} & \multicolumn{2}{c|}{CIFAR-FS} & \multicolumn{2}{c}{FC100} \\
             &          &            & 1-shot           & 5-shot         & 1-shot & 5-shot      \\ 
    \hline
    \hline
    ProtoNet~\cite{PrototypicalNetwork} & ResNet-12 & NeurIPS 2017 & - & -  &41.54±0.76 & 57.08±0.76 \\
    MetaOpt~\cite{MetaOpt} & ResNet-12 & CVPR 2019 & 72.00±0.70 & 84.20±0.50 & 41.10±0.60 & 55.50±0.60 \\
    MABAS~\cite{MABAS} & ResNet-12 & ECCV 2020 & 73.51±0.92 & 85.65±0.65 & 42.31±0.75 & 58.16±0.78 \\
    RFS~\cite{RFS} & ResNet-12 & ECCV 2020 & 73.90±0.80 & 86.90±0.50 & 44.60±0.70 & 60.90±0.60 \\
    BML~\cite{BML} & ResNet-12 & ICCV 2021 & 73.45±0.47 & 88.04±0.33 & -& - \\
    CG~\cite{CG/CNL} & ResNet-12 & AAAI 2021 & 73.00±0.70 &85.80±0.50 & -& - \\
    Meta-NVG~\cite{Mata-NVG} & ResNet-12 & ICCV 2021 & 74.63±0.91 & 86.45±0.59 & 46.40±0.81 & 61.33±0.71 \\ 
    RENet~\cite{RENet} & ResNet-12 & ICCV 2021 & 74.51±0.46 & 86.60±0.32 & - & - \\
    TPMN~\cite{TPMN} & ResNet-12 & ICCV 2021 & 75.50±0.90 & 87.20±0.60 & 46.93±0.71 & 63.26±0.74 \\
    MixFSL~\cite{MixFSL} & ResNet-12 & ICCV 2021 & -& -& 44.89±0.63 & 60.70±0.60 \\
    QSFormer~\cite{qsformer} &ResNet-12 & TCSVT 2023 & - & - & 46.51±0.26 & 61.58±0.25 \\
    LastShot~\cite{lastshot} &ResNet-12 & TPAMI 2024 & 76.76±0.21 & 87.49±0.12 & 44.08±0.18 & 59.14±0.18 \\
    \hline
    CC+rot~\cite{CC-rot} & WRN-28-10 & ICCV 2019 & 73.62±0.31 & 86.05±0.22 & -& - \\
    PSST~\cite{PSST} & WRN-28-10 & CVPR 2021 & 77.02±0.38 & 88.45±0.35 & -& - \\
    Meta-QDA~\cite{MetaQDA} & WRN-28-10 & ICCV 2021 & 75.83±0.88 & 88.79±0.75 & - & -\\
    \hline
    FewTURE~\cite{FewTrue}  & ViT-Small & NeurIPS 2022 & 76.10±0.88 & 86.14±0.64 &46.20±0.79 &63.14±0.73 \\
    % FewTURE~\cite{FewTrue}  & Swin-Tiny & 29 M & 77.76±0.81 &88.90±0.59 &47.68±0.78 &63.81±0.75 \\
    CPEA~\cite{cpea} & ViT-Small & ICCV 2023 & 77.82±0.66 & 88.98±0.45 & 47.24±0.58 & 65.02±0.60 \\
    \hline
    \name (ours) & ViT-Small & - & \bf79.81±0.67	& \bf90.81±0.67	& \bf47.79±0.61	& \bf66.69±0.58 \\
    \hline
    \end{tabular}
\end{center}

\end{table*}

\subsection{Performance Comparison} 
In accordance with established conventions in few-shot learning, we conduct experiments on four popular few-shot classification benchmarks, with the results presented in Tables~\ref{tab:miniImage} and \ref{tab:cifar}. The results indicate that our proposed method, \name, consistently achieves competitive performance compared to state-of-the-art (SOTA) approaches on both 5-way 1-shot and 5-way 5-shot tasks.

\textbf{Results on $mini$Imagenet and $tiered$Imagenet datasets}. 
Table~\ref{tab:miniImage} presents a comparison of the 1-shot and 5-shot performance of our method against baseline models on the $mini$Imagenet and $tiered$Imagenet datasets. We achieve significant improvements over existing SOTA results by leveraging both global and local features. For instance, on the $mini$Imagenet dataset, our method, \name, surpasses FewTURE by 2.02\% in the 1-shot setting and 3.49\% in the 5-shot setting. Similarly, on the $tiered$Imagenet dataset, \name outperforms FewTURE by 5.04\% and 4.28\% in the 1-shot and 5-shot settings, respectively. Compared to CPEA, which employs global features to enhance local features, our method achieves competitive results in the 1-shot setting and approximately 1\% improvement in the 5-shot setting.

Compared to FewTURE and CPEA, our method demonstrates superior accuracy while effectively utilizing both global and local features. The significant performance gap between our approach and the comparative baselines further validates the contributions of our method, which captures more comprehensive information from the data and maintains strong generalization capability.

\textbf{Results on small-resolution datasets}.
To assess the adaptability of our model, we conduct further experiments on two small-resolution datasets and compare our results with those of other methods. This approach allows us to evaluate the performance of our method across various data scenarios, ensuring a fair and comprehensive analysis.

Table~\ref{tab:cifar} presents the 1-shot and 5-shot classification performance on the small-resolution datasets CIFAR-FS and FC100. For CIFAR-FS, \name outperforms CPEA by 1.99\% in the 1-shot setting and 1.83\% in the 5-shot setting. Similarly, for FC100, our method surpasses CPEA by 0.55\% in the 1-shot setting and 1.67\% in the 5-shot setting.

These results demonstrate the effectiveness of our method across all four datasets, as it consistently achieves superior performance in all settings. The substantial improvements observed further affirm the strengths of our approach, which effectively utilizes both global and local features to enhance image representation. The efficacy of this method is evidenced by the significant performance gains achieved across diverse few-shot learning tasks.

\subsection{Ablation Study} 
In this subsection, we conduct ablation experiments to analyze the impact of each component on the performance of our method. Specifically, we focus on the 5-way setting using the $mini$Imagenet dataset and the pre-trained Vision Transformer backbone. By examining the performance variations resulting from these ablations, we gain insights into the contribution of each component to the overall effectiveness of our method. In particular, we investigate the influence of the following variations:

\begin{table}[h!]
\caption{Results of the various features utilized in metrics for few-shot classification on $mini$ImageNet.
}
\label{tab:twometric}
\begin{center}
% \scalebox{1.0}{
    \begin{tabular}{l|cc}
    \hline
    Method    &     1-shot      &      5-shot   \\ 
    \hline
    \hline
   \name(w/o local)  & 69.08±0.59  & 86.19±0.35 \\
   \name(w/o global) & 70.67±0.69  & 86.31±0.38 \\
   \name(ours) & \bf72.04±0.62  & \bf88.00±0.37 \\
    \hline
    \end{tabular}
\end{center}
\end{table}

\begin{table}[h!]
\caption{Results of the different distance functions employed in metrics for few-shot classification on $mini$ImageNet.
}
\label{tab:distance}
\begin{center}
% \scalebox{1.0}{
    \begin{tabular}{c|cc}
    \hline
    Distance Function    &     1-shot      &      5-shot   \\ 
    \hline
    \hline
   $dot$  & 68.07±0.61  & 83.43±0.65 \\
   $abs$  & 67.85±0.61  & 84.47±0.73 \\
   $cos$  & 64.06±0.69  & 75.87±0.50 \\
   $sqr$  & \bf69.08±0.59  & \bf86.19±0.35 \\
    \hline
   $wass$  & 57.85±0.71  & 66.27±0.61 \\
   $covar$  & 68.43±0.66  & 80.83±0.46 \\
   $KL$  & \bf70.67±0.69  & \bf86.31±0.38 \\
   \hline
    \end{tabular}
\end{center}
\end{table}

\begin{table}[h!]
\caption{Results of the various metric fusions for 1-shot classification on $mini$ImageNet.}
\label{tab:1-shot}
\begin{center}
\begin{tabular}{l|llll}
\hline
                       &\multicolumn{1}{c}{$wass$}  & \multicolumn{1}{c}{$covar$}        &\multicolumn{1}{c} {$KL$}          \\ 
\hline
\hline
\multicolumn{1}{l|}{$dot$} & \multicolumn{1}{c}{64.60±0.62} & \multicolumn{1}{c}{70.59±0.63} & \multicolumn{1}{l}{71.36±0.61}   \\ 
\multicolumn{1}{l|}{$abs$} & \multicolumn{1}{c}{68.16±0.65} & \multicolumn{1}{c}{70.71±0.64} & \multicolumn{1}{l}{71.49±0.62}   \\ 
\multicolumn{1}{l|}{$cos$} & \multicolumn{1}{c}{61.92±0.66} & \multicolumn{1}{c}{68.65±0.66} & \multicolumn{1}{c}{70.85±0.61}   \\ 
\multicolumn{1}{c|}{$sqr$} & \multicolumn{1}{c}{68.92±0.66} & \multicolumn{1}{c}{70.89±0.65} & \multicolumn{1}{c}{\bf72.04±0.62}   \\ 
\hline
\end{tabular}
\end{center}
\end{table}

\begin{table}[h!]
\caption{Results of the different metric fusions for 5-shot classification on $mini$ImageNet.
Results of the various metric fusions for 5-shot classification on MiniImageNet.}
\label{tab:5-shot}
\begin{center}
\begin{tabular}{l|llll}
\hline
                       &\multicolumn{1}{c}{$wass$}  & \multicolumn{1}{c}{$covar$}        &\multicolumn{1}{c} {$KL$}          \\ 
\hline
\hline
\multicolumn{1}{l|}{$dot$} & \multicolumn{1}{c}{85.75±0.38} & \multicolumn{1}{c}{87.22±0.38} & \multicolumn{1}{l}{87.27±0.36}   \\ 
\multicolumn{1}{l|}{$abs$} & \multicolumn{1}{c}{84.97±0.40} & \multicolumn{1}{c}{86.87±0.39} & \multicolumn{1}{l}{87.16±0.36}   \\ 
\multicolumn{1}{l|}{$cos$} & \multicolumn{1}{c}{84.86±0.40} & \multicolumn{1}{c}{82.41±0.47} & \multicolumn{1}{c}{86.66±0.38}   \\ 
\multicolumn{1}{c|}{$sqr$} & \multicolumn{1}{c}{86.37±0.38} & \multicolumn{1}{c}{87.57±0.37} & \multicolumn{1}{c}{\bf88.00±0.37}   \\ 
\hline
\end{tabular}
\end{center}
\end{table}

\textbf{Utilizing both global and local features}: 
To evaluate the effectiveness of our Siamese Transformer Network, we design two comparative models. In the first approach, we exclude local features, and both branches use global features to compute the similarity score. In the second approach, we exclude global features, and both branches rely solely on local features to measure the similarity score.

As shown in Table~\ref{tab:twometric}, the results clearly demonstrate the significant improvement achieved by our method, \name, compared to the baseline models. This outcome confirms the effectiveness of \name in enhancing performance for few-shot learning tasks.

\textbf{Distance functions employed in \name}: 
To identify the most suitable metric for global feature-based measurement, we evaluate four methods: dot product ($dot$), Manhattan distance ($abs$), Euclidean distance ($sqr$), and cosine similarity ($cos$). Additionally, to determine the optimal metric for local feature-based measurement, we examine three methods: 2-Wasserstein distance ($wass$), covariance metric ($covar$), and Kullback–Leibler ($KL$) divergence.

Table~\ref{tab:distance} presents the impact of different distance functions on the measurement of global and local features. The experiments demonstrate that Euclidean distance yields the most significant results for global features, while Kullback–Leibler divergence performs better for local features. Additionally, we explore various combinations of global and local feature-based metrics, as shown in Tables~\ref{tab:1-shot} and~\ref{tab:5-shot}. Our findings indicate that the results obtained from combining these two metrics generally outperform those based on a single feature metric. Notably, the fusion of Euclidean distance and KL divergence produces the best outcomes, making this our preferred choice in this study.

\begin{table}[h!]
\caption{Results of parameter sharing between two branches for few-shot classification on $mini$ImageNet. 
}
\label{tab:share}
\begin{center}
% \scalebox{1.0}{
    \begin{tabular}{c|cc}
    \hline
    Parameters Sharing    &     1-shot      &      5-shot   \\ 
    \hline
    \hline
   $\surd$  & 70.69±0.63  & 86.98±0.37 \\
   $\times$ & \bf72.04±0.62  & \bf88.00±0.37 \\
   \hline
    \end{tabular}
\end{center}
\end{table}

\textbf{Parameters between the two branch networks}: 
In our method, although the two branch networks share the same architecture, their parameters are not shared. To validate this design, we conduct comparative experiments in which the parameters are shared between the two networks, and the model loss is computed as the sum of the losses from both branch networks.

As shown in Table~\ref{tab:share}, our approach outperforms the parameter-sharing model. We speculate that this improvement is due to the interference between global feature optimization and local feature optimization when parameters are shared, which hinders the model's ability to fully leverage the strengths of each optimization. Notably, even when parameters are not shared, if both branches are evaluated using the same type of feature—either global or local—the performance is inferior to that of the parameter-sharing model that utilizes both feature types. This conclusion is supported by the comparisons between Table~\ref{tab:twometric} and Table~\ref{tab:share}. These results highlight the effectiveness of our approach in leveraging both global and local features.

\begin{table}[h!]
\caption{Results of the different fusion strategies for few-shot classification on $mini$ImageNet.
}
\label{tab:adaptive}
\begin{center}
% \scalebox{1.0}{
    \begin{tabular}{c|cc}
    \hline
    Fusion Strategy    &     1-shot      &      5-shot   \\ 
    \hline
    \hline
   Manual  & \bf72.04±0.62  & \bf88.00±0.37 \\
   Adaptive & 69.69±0.63  & 86.78±0.37 \\
   \hline
    \end{tabular}
\end{center}
\end{table}

\begin{figure}[t] %插入图片
\centering %图片居中
\begin{tikzpicture}[scale=0.8] %tikz图片
\begin{axis}[
    xlabel=$\alpha$, %横坐标名
    % \vspace{3mm}
    ylabel=Accuracy on validation set $(\%)$, %纵坐标名
    tick align=inside, %刻度在外显式
    legend style={at={(0.90,0.55))},anchor=south east} %图例在图下方显示
    ]
%第一条线，mark是折线标示形状
\addplot[smooth,blue, line width = 1.2pt] plot coordinates { 
    (0.1,69.22)
    (0.2,70.24)
    (0.3,70.77)
    (0.4,71.22)
    (0.5,71.60)
    (0.6,71.91)
    (0.7,72.04)
    (0.8,72.00)
    (0.9,71.76)
};
\addlegendentry{$1-shot$}

\addplot[smooth,black, line width = 1.2pt] plot coordinates { 
    (0.1,86.85)
    (0.2,87.28)
    (0.3,87.59)
    (0.4,87.84)
    (0.5,87.94)
    (0.6,88.00)
    (0.7,87.84)
    (0.8,87.62)
    (0.9,87.14)
};
\addlegendentry{$5-shot$}
\end{axis}
\end{tikzpicture}
\caption{Results of different weights in the fusion strategy for few-shot classification on $mini$ImageNet.}
\label{fig:alpha}
\end{figure}
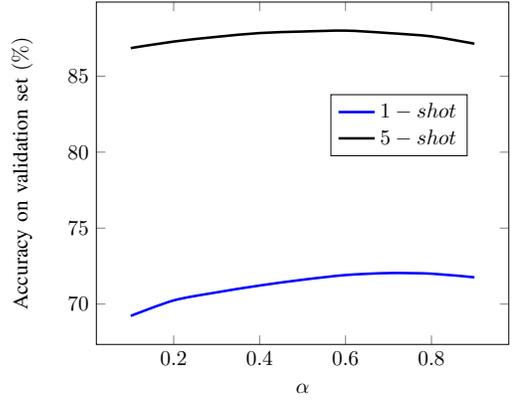

\textbf{Manual fusion strategy vs. adaptive fusion strategy}: 
In this paper, we adopt a manually designed fusion strategy. To validate this design, we also implement an adaptive fusion strategy for comparison. In the adaptive fusion strategy, we utilize a learnable 2-dimensional weight vector \( w = [\omega_1; \omega_2] \) to perform the fusion. Thus, the final fusion similarity between a query \( Q \) and a class \( S \) can be defined as follows:

\begin{align}\label{e14}
    D(Q, S) = -\omega_1 \cdot D_{KL}(Q || S) - \omega_2 \cdot D_{Ed}(Q, S).
\end{align}

Specifically, we concatenate these two metric vectors to form a 10-dimensional vector. Subsequently, we apply a 1-dimensional convolutional layer with a kernel size of \(1 \times 1\) and a dilation rate of 5. This process allows us to obtain a weighted 5-dimensional similarity vector by learning a 2-dimensional weight \(\omega\). Additionally, a batch normalization layer is added before the convolutional layer to balance the scale of the similarities from both components. Finally, a non-parametric nearest neighbor classifier is employed to derive the final classification result.

As shown in Table~\ref{tab:adaptive}, our approach outperforms the model utilizing the adaptive fusion strategy. We hypothesize that while the adaptive method can learn to adjust, it may lead to overfitting on the training data, resulting in underperformance on unseen test data. In contrast, using fixed weights reduces model complexity, thereby mitigating overfitting and enhancing the model's generalization capability.

\begin{figure*}[t]
\begin{center}
  \includegraphics[width=1.0\linewidth]{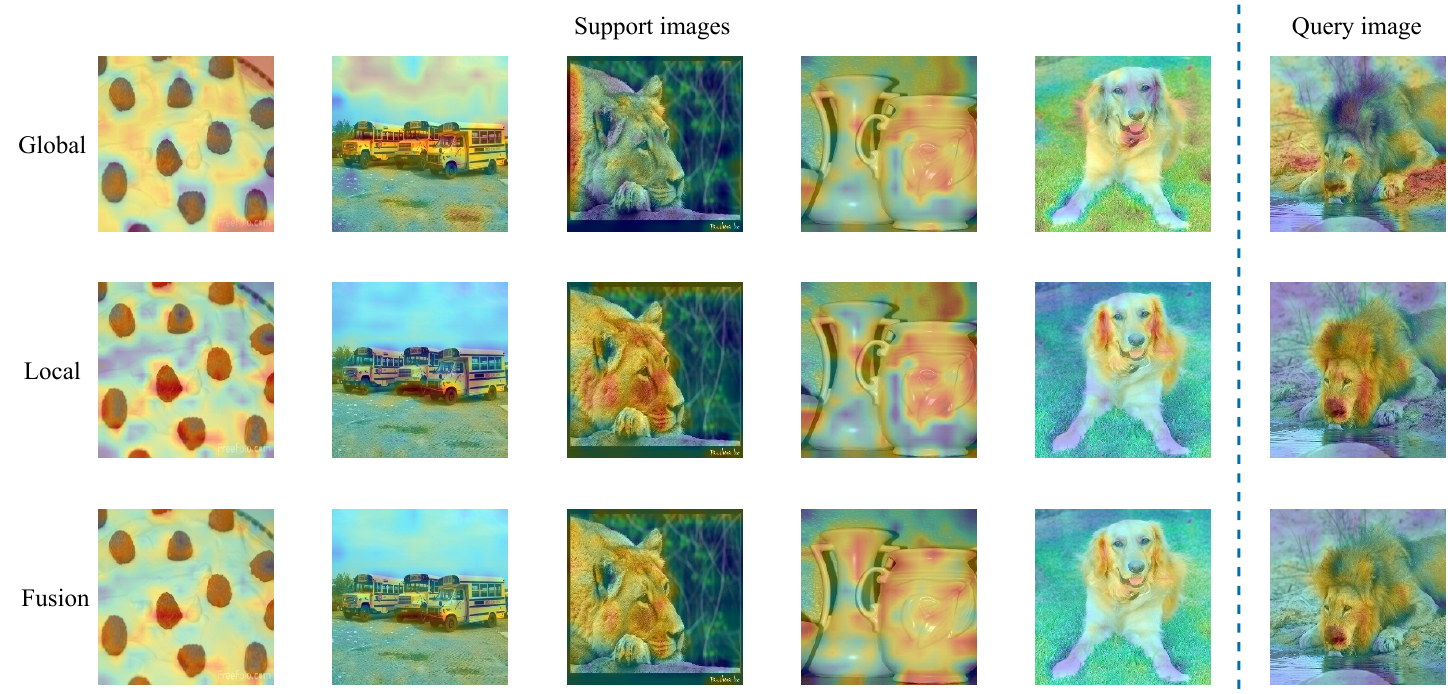}
\end{center}
   \caption{ 
   Attention map visualization. %The color degree indicates the degree of correlation between a local region and the global information. The darker the red, the higher the correlation, and the darker the blue, the lower the correlation. With global and local information fusion, our proposed method decreases the weights placed on the semantics that are irrelevant to the global information.
  The color intensity indicates the level of correlation between a local region and global information. Darker red shades signify higher correlation, while darker blue shades indicate lower correlation. By fusing global and local information, our proposed method reduces the weights assigned to semantics that are irrelevant to the global context.}
\label{fig:fusion}
\end{figure*}  

% \textbf{Training strategy}:
\begin{table}[h!]
\caption{
Results of normalization in the fusion strategy for few-shot classification on $mini$ImageNet.
}
\label{tab:norm}
\begin{center}
% \scalebox{1.0}{
    \begin{tabular}{c|cc}
    \hline
    Normalization    &     1-shot      &      5-shot   \\ 
    \hline
    \hline
   $\surd$  & \bf72.04±0.62  & \bf88.00±0.37 \\
   $\times$ & 70.74±0.69  & 86.69±0.38 \\
   \hline
    \end{tabular}
\end{center}
\end{table}

\textbf{Fusion of dual metrics with normalization}:
In our method, we first normalize the two metrics before fusing them. To validate the necessity of normalization, we fused the original results of the two metrics without normalization. The experimental results, as shown in Table~\ref{tab:norm}, indicate a significant decrease in model performance when normalization is omitted, highlighting the importance of this step. 
Additionally, we examined the impact of different weights on evaluation performance, with results presented in Figure~\ref{fig:alpha}. The findings reveal that placing greater emphasis on the measurement results of the feature type with superior performance leads to improved outcomes. However, there is a tipping point: for the $mini$ImageNet dataset in the 1-shot setting, the optimal value of \(\alpha\) is 0.7, while in the 5-shot setting, it decreases to 0.6.

\subsection{Visualization Analysis} 
To evaluate the proposed approach, we randomly select a task from the $mini$ImageNet dataset to demonstrate the changes in the model's attention map. As shown in Figure~\ref{fig:fusion}, after applying the feature fusion process, the model exhibits an improved ability to focus on the key areas of the images. These results further validate the effectiveness of our approach, indicating that by fusing global and local features from a single image, the resulting query and support features become more distinguishable.

%% file: 05Conclusion.tex
In few-shot learning based on metric learning, feature adaptation modules are typically designed to enhance global feature representation or align semantically related local regions. Existing approaches often learn the similarity between support and query pairs using either global features or local features, neglecting 
%the combined use of both. 
the use of their combination.
To address this limitation, we propose a Siamese Transformer Network (STN) framework that employs a dual-branch architecture to extract global and local features separately. To effectively leverage these features, we utilize Euclidean distance for global features and Kullback-Leibler divergence for local features. Based on these measurements, we develop a simple additive fusion strategy that normalizes and weights the two metrics to produce a final similarity evaluation.
This study %presents a novel perspective and an
introduces a simple yet 
effective method for metric-based few-shot learning by exploiting the strengths of both global and local features. 
%Our approach is simple yet very effective. 
Quantitative experiments on four benchmark datasets validate its effectiveness. %, demonstrating that it 
Our method achieves state-of-the-art performance while maintaining a straightforward and uncomplicated structure. This simplicity not only ensures ease of implementation but also highlights the robustness and reliability of our method.

%% file: 00Main.bbl
\begin{thebibliography}{10}
\providecommand{\url}[1]{#1}
\csname url@samestyle\endcsname
\providecommand{\newblock}{\relax}
\providecommand{\bibinfo}[2]{#2}
\providecommand{\BIBentrySTDinterwordspacing}{\spaceskip=0pt\relax}
\providecommand{\BIBentryALTinterwordstretchfactor}{4}
\providecommand{\BIBentryALTinterwordspacing}{\spaceskip=\fontdimen2\font plus
\BIBentryALTinterwordstretchfactor\fontdimen3\font minus \fontdimen4\font\relax}
\providecommand{\BIBforeignlanguage}[2]{{%
\expandafter\ifx\csname l@#1\endcsname\relax
\typeout{** WARNING: IEEEtran.bst: No hyphenation pattern has been}%
\typeout{** loaded for the language `#1'. Using the pattern for}%
\typeout{** the default language instead.}%
\else
\language=\csname l@#1\endcsname
\fi
#2}}
\providecommand{\BIBdecl}{\relax}
\BIBdecl

\bibitem{siamese}
J.~Bromley, J.~W. Bentz, L.~Bottou, I.~Guyon, Y.~LeCun, C.~Moore, E.~S{\"{a}}ckinger, and R.~Shah, ``Signature verification using {A} "siamese" time delay neural network,'' \emph{Int. J. Pattern Recognit. Artif. Intell.}, vol.~7, no.~4, pp. 669--688, 1993.

\bibitem{one-shot}
B.~M. Lake, R.~Salakhutdinov, J.~Gross, and J.~B. Tenenbaum, ``One shot learning of simple visual concepts,'' in \emph{Proceedings of the 33th Annual Meeting of the Cognitive Science Society}, 2011.

\bibitem{PrototypicalNetwork}
J.~Snell, K.~Swersky, and R.~S. Zemel, ``Prototypical networks for few-shot learning,'' in \emph{Advances in Neural Information Processing Systems}, 2017.

\bibitem{hri}
X.~Li, J.~Chen, H.~Zhang, Y.~Cho, S.~H. Hwang, Z.~Gao, and G.~Yang, ``Hierarchical relational inference for few-shot learning in 3d left atrial segmentation,'' \emph{IEEE Transactions on Emerging Topics in Computational Intelligence}, pp. 1--16, 2024.

\bibitem{joc}
S.~Fu, Q.~Peng, X.~Wang, Y.~He, W.~Qiu, B.~Zou, D.~Xu, X.-Y. Jing, and X.~You, ``Jointly optimized classifiers for few-shot class-incremental learning,'' \emph{IEEE Transactions on Emerging Topics in Computational Intelligence}, pp. 1--11, 2024.

\bibitem{RelationNetwork}
F.~Sung, Y.~Yang, L.~Zhang, T.~Xiang, P.~H.~S. Torr, and T.~M. Hospedales, ``Learning to compare: Relation network for few-shot learning,'' in \emph{{IEEE/CVF} Conference on Computer Vision and Pattern Recognition}, 2018.

\bibitem{Meta-Baseline}
Y.~Chen, Z.~Liu, H.~Xu, T.~Darrell, and X.~Wang, ``Meta-baseline: Exploring simple meta-learning for few-shot learning,'' in \emph{{IEEE/CVF} International Conference on Computer Vision}, 2021.

\bibitem{MatchingNetwork}
O.~Vinyals, C.~Blundell, T.~Lillicrap, k.~kavukcuoglu, and D.~Wierstra, ``Matching networks for one shot learning,'' in \emph{Advances in Neural Information Processing Systems}, 2016.

\bibitem{msml}
W.~Jiang, K.~Huang, J.~Geng, and X.~Deng, ``Multi-scale metric learning for few-shot learning,'' \emph{IEEE Transactions on Circuits and Systems for Video Technology}, vol.~31, no.~3, pp. 1091--1102, 2021.

\bibitem{deepemd}
C.~Zhang, Y.~Cai, G.~Lin, and C.~Shen, ``Deepemd: Few-shot image classification with differentiable earth mover's distance and structured classifiers,'' in \emph{{IEEE/CVF} Conference on Computer Vision and Pattern Recognition}, 2020.

\bibitem{CTM}
H.~Li, D.~Eigen, S.~Dodge, M.~Zeiler, and X.~Wang, ``Finding task-relevant features for few-shot learning by category traversal,'' in \emph{{IEEE} Conference on Computer Vision and Pattern Recognition, {CVPR} 2019, Long Beach, CA, USA, June 16-20, 2019}.\hskip 1em plus 0.5em minus 0.4em\relax Computer Vision Foundation / {IEEE}, 2019, pp. 1--10.

\bibitem{FEAT}
H.~Ye, H.~Hu, D.~Zhan, and F.~Sha, ``Few-shot learning via embedding adaptation with set-to-set functions,'' in \emph{{IEEE/CVF} Conference on Computer Vision and Pattern Recognition}, 2020.

\bibitem{MAML}
C.~Finn, P.~Abbeel, and S.~Levine, ``Model-agnostic meta-learning for fast adaptation of deep networks,'' in \emph{Proceedings of the 34th International Conference on Machine Learning}, 2017.

\bibitem{ANIL}
A.~Raghu, M.~Raghu, S.~Bengio, and O.~Vinyals, ``Rapid learning or feature reuse? towards understanding the effectiveness of {MAML},'' in \emph{8th International Conference on Learning Representations}, 2020.

\bibitem{BOIL}
J.~Oh, H.~Yoo, C.~Kim, and S.~Yun, ``{BOIL:} towards representation change for few-shot learning,'' in \emph{9th International Conference on Learning Representations}, 2021.

\bibitem{LEO}
A.~A. Rusu, D.~Rao, J.~Sygnowski, O.~Vinyals, R.~Pascanu, S.~Osindero, and R.~Hadsell, ``Meta-learning with latent embedding optimization,'' in \emph{7th International Conference on Learning Representations, {ICLR} 2019, New Orleans, LA, USA, May 6-9, 2019}.\hskip 1em plus 0.5em minus 0.4em\relax OpenReview.net, 2019.

\bibitem{dynamic}
S.~Gidaris and N.~Komodakis, ``Dynamic few-shot visual learning without forgetting,'' in \emph{{IEEE/CVF} Conference on Computer Vision and Pattern Recognition}, 2018.

\bibitem{baseline++}
W.~Chen, Y.~Liu, Z.~Kira, Y.~F. Wang, and J.~Huang, ``A closer look at few-shot classification,'' in \emph{7th International Conference on Learning Representations, {ICLR} 2019, New Orleans, LA, USA, May 6-9, 2019}.\hskip 1em plus 0.5em minus 0.4em\relax OpenReview.net, 2019.

\bibitem{RFS}
Y.~Tian, Y.~Wang, D.~Krishnan, J.~B. Tenenbaum, and P.~Isola, ``Rethinking few-shot image classification: {A} good embedding is all you need?'' in \emph{Computer Vision - {ECCV} 2020 - 16th European Conference, Glasgow, UK, August 23-28, 2020, Proceedings, Part {XIV}}.\hskip 1em plus 0.5em minus 0.4em\relax Springer, 2020, pp. 266--282.

\bibitem{P>M>F}
S.~X. Hu, D.~Li, J.~St{\"{u}}hmer, M.~Kim, and T.~M. Hospedales, ``Pushing the limits of simple pipelines for few-shot learning: External data and fine-tuning make a difference,'' in \emph{{IEEE/CVF} Conference on Computer Vision and Pattern Recognition}, 2022.

\bibitem{HCTransformer}
Y.~He, W.~Liang, D.~Zhao, H.~Zhou, W.~Ge, Y.~Yu, and W.~Zhang, ``Attribute surrogates learning and spectral tokens pooling in transformers for few-shot learning,'' in \emph{{IEEE/CVF} Conference on Computer Vision and Pattern Recognition, {CVPR} 2022, New Orleans, LA, USA, June 18-24, 2022}.\hskip 1em plus 0.5em minus 0.4em\relax {IEEE}, 2022, pp. 9109--9119.

\bibitem{covamnet}
W.~Li, J.~Xu, J.~Huo, L.~Wang, Y.~Gao, and J.~Luo, ``Distribution consistency based covariance metric networks for few-shot learning,'' in \emph{The Thirty-Third {AAAI} Conference on Artificial Intelligence, {AAAI} 2019, The Thirty-First Innovative Applications of Artificial Intelligence Conference, {IAAI} 2019, The Ninth {AAAI} Symposium on Educational Advances in Artificial Intelligence, {EAAI} 2019, Honolulu, Hawaii, USA, January 27 - February 1, 2019}.\hskip 1em plus 0.5em minus 0.4em\relax {AAAI} Press, 2019, pp. 8642--8649.

\bibitem{DN4}
W.~Li, L.~Wang, J.~Xu, J.~Huo, Y.~Gao, and J.~Luo, ``Revisiting local descriptor based image-to-class measure for few-shot learning,'' in \emph{{IEEE} Conference on Computer Vision and Pattern Recognition, {CVPR} 2019, Long Beach, CA, USA, June 16-20, 2019}.\hskip 1em plus 0.5em minus 0.4em\relax Computer Vision Foundation / {IEEE}, 2019, pp. 7260--7268.

\bibitem{adm}
W.~Li, L.~Wang, J.~Huo, Y.~Shi, Y.~Gao, and J.~Luo, ``Asymmetric distribution measure for few-shot learning,'' in \emph{Proceedings of the Twenty-Ninth International Joint Conference on Artificial Intelligence, {IJCAI} 2020}, C.~Bessiere, Ed.\hskip 1em plus 0.5em minus 0.4em\relax ijcai.org, 2020, pp. 2957--2963.

\bibitem{CAN}
R.~Hou, H.~Chang, B.~Ma, S.~Shan, and X.~Chen, ``Cross attention network for few-shot classification,'' in \emph{Advances in Neural Information Processing Systems, December 8-14, 2019, Vancouver, BC, Canada}, H.~M. Wallach, H.~Larochelle, A.~Beygelzimer, F.~d'Alch{\'{e}}{-}Buc, E.~B. Fox, and R.~Garnett, Eds., 2019, pp. 4005--4016.

\bibitem{CTX}
C.~Doersch, A.~Gupta, and A.~Zisserman, ``Crosstransformers: spatially-aware few-shot transfer,'' in \emph{Advances in Neural Information Processing Systems 33: Annual Conference on Neural Information Processing Systems 2020, NeurIPS 2020, December 6-12, 2020, virtual}, H.~Larochelle, M.~Ranzato, R.~Hadsell, M.~Balcan, and H.~Lin, Eds., 2020.

\bibitem{ATL-Net}
C.~Dong, W.~Li, J.~Huo, Z.~Gu, and Y.~Gao, ``Learning task-aware local representations for few-shot learning,'' in \emph{Proceedings of the Twenty-Ninth International Joint Conference on Artificial Intelligence}, 2020.

\bibitem{RENet}
D.~Kang, H.~Kwon, J.~Min, and M.~Cho, ``Relational embedding for few-shot classification,'' in \emph{2021 {IEEE/CVF} International Conference on Computer Vision, {ICCV} 2021, Montreal, QC, Canada, October 10-17, 2021}.\hskip 1em plus 0.5em minus 0.4em\relax {IEEE}, 2021, pp. 8802--8813.

\bibitem{FewTrue}
M.~Hiller, R.~Ma, M.~Harandi, and T.~Drummond, ``Rethinking generalization in few-shot classification,'' in \emph{Advances in Neural Information Processing Systems}, 2022.

\bibitem{cpea}
F.~Hao, F.~He, L.~Liu, F.~Wu, D.~Tao, and J.~Cheng, ``Class-aware patch embedding adaptation for few-shot image classification,'' in \emph{{IEEE/CVF} International Conference on Computer Vision, {ICCV} 2023, Paris, France, October 1-6, 2023}.\hskip 1em plus 0.5em minus 0.4em\relax {IEEE}, 2023, pp. 18\,859--18\,869.

\bibitem{mixer}
J.~Cheng, F.~Hao, F.~He, L.~Liu, and Q.~Zhang, ``Mixer-based semantic spread for few-shot learning,'' \emph{{IEEE} Trans. Multim.}, vol.~25, pp. 191--202, 2023.

\bibitem{saml}
F.~Hao, F.~He, J.~Cheng, L.~Wang, J.~Cao, and D.~Tao, ``Collect and select: Semantic alignment metric learning for few-shot learning,'' in \emph{2019 {IEEE/CVF} International Conference on Computer Vision, {ICCV} 2019, Seoul, Korea (South), October 27 - November 2, 2019}.\hskip 1em plus 0.5em minus 0.4em\relax {IEEE}, 2019, pp. 8459--8468.

\bibitem{2015siamese}
G.~Koch, R.~Zemel, R.~Salakhutdinov \emph{et~al.}, ``Siamese neural networks for one-shot image recognition,'' in \emph{ICML Deep Learning Workshop}, 2015.

\bibitem{siam}
L.~H. Hughes, M.~Schmitt, L.~Mou, Y.~Wang, and X.~X. Zhu, ``Identifying corresponding patches in {SAR} and optical images with a pseudo-siamese {CNN},'' \emph{{IEEE} Geosci. Remote. Sens. Lett.}, vol.~15, no.~5, pp. 784--788, 2018.

\bibitem{ViT}
A.~Dosovitskiy, L.~Beyer, A.~Kolesnikov, D.~Weissenborn, X.~Zhai, T.~Unterthiner, M.~Dehghani, M.~Minderer, G.~Heigold, S.~Gelly, J.~Uszkoreit, and N.~Houlsby, ``An image is worth 16x16 words: Transformers for image recognition at scale,'' in \emph{9th International Conference on Learning Representations, {ICLR} 2021, Virtual Event, Austria, May 3-7, 2021}.\hskip 1em plus 0.5em minus 0.4em\relax OpenReview.net, 2021.

\bibitem{kl}
J.~C. Duchi, ``Derivations for linear algebra and optimization,'' 2016.

\bibitem{MiniImageNet}
S.~Ravi and H.~Larochelle, ``Optimization as a model for few-shot learning,'' in \emph{5th International Conference on Learning Representations}, 2017.

\bibitem{tieredImageNet}
M.~Ren, E.~Triantafillou, S.~Ravi, J.~Snell, K.~Swersky, J.~B. Tenenbaum, H.~Larochelle, and R.~S. Zemel, ``Meta-learning for semi-supervised few-shot classification,'' in \emph{6th International Conference on Learning Representations}, 2018.

\bibitem{cifar-fs}
L.~Bertinetto, J.~F. Henriques, P.~H.~S. Torr, and A.~Vedaldi, ``Meta-learning with differentiable closed-form solvers,'' in \emph{7th International Conference on Learning Representations, {ICLR} 2019, New Orleans, LA, USA, May 6-9, 2019}.\hskip 1em plus 0.5em minus 0.4em\relax OpenReview.net, 2019.

\bibitem{fc100}
B.~N. Oreshkin, P.~R. L{\'{o}}pez, and A.~Lacoste, ``{TADAM:} task dependent adaptive metric for improved few-shot learning,'' in \emph{Advances in Neural Information Processing Systems, December 3-8, 2018, Montr{\'{e}}al, Canada}, S.~Bengio, H.~M. Wallach, H.~Larochelle, K.~Grauman, N.~Cesa{-}Bianchi, and R.~Garnett, Eds., 2018, pp. 719--729.

\bibitem{ImageNet}
A.~Krizhevsky, I.~Sutskever, and G.~E. Hinton, ``Imagenet classification with deep convolutional neural networks,'' in \emph{Advances in Neural Information Processing Systems}, 2012.

\bibitem{IEPT}
M.~Zhang, J.~Zhang, Z.~Lu, T.~Xiang, M.~Ding, and S.~Huang, ``{IEPT:} instance-level and episode-level pretext tasks for few-shot learning,'' in \emph{9th International Conference on Learning Representations, {ICLR} 2021, Virtual Event, Austria, May 3-7, 2021}.\hskip 1em plus 0.5em minus 0.4em\relax OpenReview.net, 2021.

\bibitem{MELR}
N.~Fei, Z.~Lu, T.~Xiang, and S.~Huang, ``{MELR:} meta-learning via modeling episode-level relationships for few-shot learning,'' in \emph{9th International Conference on Learning Representations, {ICLR} 2021, Virtual Event, Austria, May 3-7, 2021}.\hskip 1em plus 0.5em minus 0.4em\relax OpenReview.net, 2021.

\bibitem{FRN}
D.~Wertheimer, L.~Tang, and B.~Hariharan, ``Few-shot classification with feature map reconstruction networks,'' in \emph{{IEEE/CVF} Conference on Computer Vision and Pattern Recognition}, 2021.

\bibitem{CG/CNL}
J.~Zhao, Y.~Yang, X.~Lin, J.~Yang, and L.~He, ``Looking wider for better adaptive representation in few-shot learning,'' in \emph{AAAI Conference on Artificial Intelligence}, 2021.

\bibitem{DMF}
C.~Xu, Y.~Fu, C.~Liu, C.~Wang, J.~Li, F.~Huang, L.~Zhang, and X.~Xue, ``Learning dynamic alignment via meta-filter for few-shot learning,'' in \emph{{IEEE} Conference on Computer Vision and Pattern Recognition, {CVPR} 2021, virtual, June 19-25, 2021}.\hskip 1em plus 0.5em minus 0.4em\relax Computer Vision Foundation / {IEEE}, 2021, pp. 5182--5191.

\bibitem{InfoPatch}
C.~Liu, Y.~Fu, C.~Xu, S.~Yang, J.~Li, C.~Wang, and L.~Zhang, ``Learning a few-shot embedding model with contrastive learning,'' in \emph{Thirty-Fifth {AAAI} Conference on Artificial Intelligence, {AAAI} 2021}.\hskip 1em plus 0.5em minus 0.4em\relax {AAAI} Press, 2021, pp. 8635--8643.

\bibitem{BML}
Z.~Zhou, X.~Qiu, J.~Xie, J.~Wu, and C.~Zhang, ``Binocular mutual learning for improving few-shot classification,'' in \emph{2021 {IEEE/CVF} International Conference on Computer Vision, {ICCV} 2021, Montreal, QC, Canada, October 10-17, 2021}.\hskip 1em plus 0.5em minus 0.4em\relax {IEEE}, 2021, pp. 8382--8391.

\bibitem{Mata-NVG}
C.~Zhang, H.~Ding, G.~Lin, R.~Li, C.~Wang, and C.~Shen, ``Meta navigator: Search for a good adaptation policy for few-shot learning,'' in \emph{2021 {IEEE/CVF} International Conference on Computer Vision, {ICCV} 2021, Montreal, QC, Canada, October 10-17, 2021}.\hskip 1em plus 0.5em minus 0.4em\relax {IEEE}, 2021, pp. 9415--9424.

\bibitem{PAL}
J.~Ma, H.~Xie, G.~Han, S.~Chang, A.~Galstyan, and W.~Abd{-}Almageed, ``Partner-assisted learning for few-shot image classification,'' in \emph{2021 {IEEE/CVF} International Conference on Computer Vision, {ICCV} 2021, Montreal, QC, Canada, October 10-17, 2021}.\hskip 1em plus 0.5em minus 0.4em\relax {IEEE}, 2021, pp. 10\,553--10\,562.

\bibitem{COSOC}
X.~Luo, L.~Wei, L.~Wen, J.~Yang, L.~Xie, Z.~Xu, and Q.~Tian, ``Rectifying the shortcut learning of background for few-shot learning,'' in \emph{Advances in Neural Information Processing Systems, December 6-14, 2021, virtual}, M.~Ranzato, A.~Beygelzimer, Y.~N. Dauphin, P.~Liang, and J.~W. Vaughan, Eds., 2021, pp. 13\,073--13\,085.

\bibitem{Meta-DeepBDC}
J.~Xie, F.~Long, J.~Lv, Q.~Wang, and P.~Li, ``Joint distribution matters: Deep brownian distance covariance for few-shot classification,'' in \emph{{IEEE/CVF} Conference on Computer Vision and Pattern Recognition, {CVPR} 2022, New Orleans, LA, USA, June 18-24, 2022}.\hskip 1em plus 0.5em minus 0.4em\relax {IEEE}, 2022, pp. 7962--7971.

\bibitem{qsformer}
X.~Wang, X.~Wang, B.~Jiang, and B.~Luo, ``Few-shot learning meets transformer: Unified query-support transformers for few-shot classification,'' \emph{IEEE Transactions on Circuits and Systems for Video Technology}, vol.~33, no.~12, pp. 7789--7802, 2023.

\bibitem{lastshot}
H.~Ye, L.~Ming, D.~Zhan, and W.~Chao, ``Few-shot learning with a strong teacher,'' \emph{{IEEE} Trans. Pattern Anal. Mach. Intell.}, vol.~46, no.~3, pp. 1425--1440, 2024.

\bibitem{CC-rot}
S.~Gidaris, A.~Bursuc, N.~Komodakis, P.~P{\'{e}}rez, and M.~Cord, ``Boosting few-shot visual learning with self-supervision,'' in \emph{2019 {IEEE/CVF} International Conference on Computer Vision, {ICCV} 2019, Seoul, Korea (South), October 27 - November 2, 2019}.\hskip 1em plus 0.5em minus 0.4em\relax {IEEE}, 2019, pp. 8058--8067.

\bibitem{MetaQDA}
X.~Zhang, D.~Meng, H.~Gouk, and T.~M. Hospedales, ``Shallow bayesian meta learning for real-world few-shot recognition,'' in \emph{2021 {IEEE/CVF} International Conference on Computer Vision, {ICCV} 2021, Montreal, QC, Canada, October 10-17, 2021}.\hskip 1em plus 0.5em minus 0.4em\relax {IEEE}, 2021, pp. 631--640.

\bibitem{OM}
G.~Qi, H.~Yu, Z.~Lu, and S.~Li, ``Transductive few-shot classification on the oblique manifold,'' in \emph{2021 {IEEE/CVF} International Conference on Computer Vision, {ICCV} 2021, Montreal, QC, Canada, October 10-17, 2021}.\hskip 1em plus 0.5em minus 0.4em\relax {IEEE}, 2021, pp. 8392--8402.

\bibitem{MetaOpt}
K.~Lee, S.~Maji, A.~Ravichandran, and S.~Soatto, ``Meta-learning with differentiable convex optimization,'' in \emph{{IEEE} Conference on Computer Vision and Pattern Recognition, {CVPR} 2019, Long Beach, CA, USA, June 16-20, 2019}.\hskip 1em plus 0.5em minus 0.4em\relax Computer Vision Foundation / {IEEE}, 2019, pp. 10\,657--10\,665.

\bibitem{MABAS}
J.~Kim, H.~Kim, and G.~Kim, ``Model-agnostic boundary-adversarial sampling for test-time generalization in few-shot learning,'' in \emph{Computer Vision - {ECCV} 2020 - 16th European Conference, Glasgow, UK, August 23-28, 2020, Proceedings, Part {I}}.\hskip 1em plus 0.5em minus 0.4em\relax Springer, 2020, pp. 599--617.

\bibitem{TPMN}
J.~Wu, T.~Zhang, Y.~Zhang, and F.~Wu, ``Task-aware part mining network for few-shot learning,'' in \emph{2021 {IEEE/CVF} International Conference on Computer Vision, {ICCV} 2021, Montreal, QC, Canada, October 10-17, 2021}.\hskip 1em plus 0.5em minus 0.4em\relax {IEEE}, 2021, pp. 8413--8422.

\bibitem{MixFSL}
A.~Afrasiyabi, J.~Lalonde, and C.~Gagn{\'{e}}, ``Mixture-based feature space learning for few-shot image classification,'' in \emph{2021 {IEEE/CVF} International Conference on Computer Vision, {ICCV} 2021, Montreal, QC, Canada, October 10-17, 2021}.\hskip 1em plus 0.5em minus 0.4em\relax {IEEE}, 2021, pp. 9021--9031.

\bibitem{PSST}
Z.~Chen, J.~Ge, H.~Zhan, S.~Huang, and D.~Wang, ``Pareto self-supervised training for few-shot learning,'' in \emph{{IEEE} Conference on Computer Vision and Pattern Recognition, {CVPR} 2021, virtual, June 19-25, 2021}.\hskip 1em plus 0.5em minus 0.4em\relax Computer Vision Foundation / {IEEE}, 2021, pp. 13\,663--13\,672.

\end{thebibliography}
